\documentclass[11pt]{article}
\usepackage{graphicx} % Allows including images
\usepackage{caption}
\usepackage{amsmath}
\usepackage[T1]{fontenc}
\usepackage[utf8]{inputenc}
\usepackage{authblk}
\usepackage{hyperref}
\usepackage{bbm}
\usepackage{amsthm}
\usepackage{multirow}
\DeclareMathOperator*{\argmin}{argmin} 
\usepackage{float}
\usepackage{mathtools}
\usepackage{amsfonts}
\usepackage{nicematrix}

\usepackage{cite}

\title{Coarse- and fine-scale geometric information content of Multiclass Classification and implied Data-driven Intelligence}
\author[1]{Fushing Hsieh}
\author[1]{Xiaodong Wang}
\affil[1]{Department of Statistics, University of California, Davis.}

\date{}

\begin{document}

\maketitle

\section*{abstract}
Under any Multiclass Classification (MCC) setting defined by a collection of labeled point-cloud specified by a feature-set, we extract only stochastic partial orderings from all possible triplets of point-cloud without explicitly measuring the three cloud-to-cloud distances. We demonstrate that such a collective of partial ordering can efficiently compute a label embedding tree geometry on the Label-space. This tree in turn gives rise to a predictive graph, or a network with precisely weighted linkages. Such two multiscale geometries are taken as the coarse scale information content of MCC. They indeed jointly shed lights on explainable knowledge on why and how labeling comes about and facilitates error-free prediction with potential multiple candidate labels supported by data. For revealing within-label heterogeneity, we further undergo labeling naturally found clusters within each point-cloud, and likewise derive multiscale geometry as its fine-scale information content contained in data. This fine-scale endeavor shows that our computational proposal is indeed scalable to a MCC setting having a large label-space. Overall the computed multiscale collective of data-driven patterns and knowledge will serve as a basis for constructing visible and explainable subject matter intelligence regarding the system of interest. 

\section{Introduction}
Nowadays Machine Learning (M.L.) based Artificial Intelligence (A.I.) researches are by-and-large charged to endow machines with various human's semantic categorizing capabilities \cite{1}. Given that human experts hardly make semantic categorizing mistakes, should machine also help to explain: How and Why, to human? We demonstrate that possible answers are computational and visible under any Multiclass Classification (MCC) setting. The keys are: first compute the pertinent information content without artificial structure; secondly, graphically display such information content via multiscale geometries, such as a tree, a network or both, to concisely organize and deliver pattern-based knowledge or intelligence contained in data to human attentions.

Multiclass Classification is one major topic \cite{5,6,7,8,9} of associating visual images or text articles with semantic concepts \cite{2,3,4}. Its two popular techniques: flat and hierarchical, are prone to make mistakes \cite{10,11,12}. Since a machine is primarily forced to assign a single candidate label toward a prediction. No less, no more. Such a forceful decision-making to a great extent ignores the available amount of information supported by data. With such kind of M.L. in the heart of A.I., it is beyond reasonable doubt that A.I. is bound to generate fundamental social and academic issues in the foreseeable future, if its error-prone propensity is not well harnessed in time.

If completely error-free A.I. is not possible at current state of technology, then at least it should tell us its decision-making trajectory leading up to every right or wrong decision. It is in the same sense as the recommended fourth rule of robotics: ``a robot or any intelligent machine-must be able to explain itself to humans'' to be added to Asimov's famous three. Since we need to see why, how and where errors occur in hope of knowing what causes, and even figuring out how to fix it.

Such a quality prerequisite on A.I. and M.L. is also coherent with concurrent requirements put forth by many governments around the world: Transparent explanation upon each A.I. based decision is required. Now it is a critical time point to think about how to coherently build and display data's authentic information content that can afford the making of explainable error-free decisions. So such information content with pertinent graphic display can be turned into Data-driven Intelligence. In this paper, we specifically demonstrate Data-driven Intelligence for Multiclass Classification. This choice of M.L. topic is in part due to that classification is human's primary way of acquiring intelligence, and also in part due to its fundamental importance in science and industry.

On the road to Data-driven Intelligence, we begin by asking the following three simple questions. First, the naive one is: Where is relevant information in data? Secondly, what metric geometry is suitable to represent such information content? Finally, how to make perfect, or at least nearly perfect empirical inference or predictive decision-making? We address these three non-hypothetical questions thoroughly based on model-free unsupervised M.L. Here we explicitly show the nature of information content under Multiclass Classification as: multiscale heterogeneity. Such information heterogeneity can be rather intertwined and opaque when its three data-scales: numbers of label, feature and instance, are all big.

The paper is organized as follows. In section 2 we describe the background and related work of MCC. In section 3 we develop a new label-embedding tree constructed via partial ordering and a classification schedule. In section 4 we illustrate a tree-decent procedure with early stop and represent the error flow. In section 5 we explore the heterogeneity embedded within labels. 

\section{Multiclass Classification}
A generic Multiclass Classification (MCC) setting has three data scales: the number of label $L$, the number of feature $K$ and total number of subjects $N$. Each label specifies a data-cloud. A data-cloud is an ensemble of subjects. Each subject is identified by a vector of $K$ feature measurements. The complexity of data and its information content under any MCC setting is critically subject to $L$, $K$ and $N$. The goal of Multiclass Classification is to seek for the principles or intelligence that can explain label-to-feature linkages. Such linkages are intrinsically heterogeneous as being blurred by varying degrees of mixing among diverse groups within the space of labeled data-clouds. Since such data mixing patterns are likely rather convoluted and intertwined, so the overall complexity of information content must be multiscale in nature.

Specifically speaking, its global scale is referred to which label's point-cloud is close to which, but far away from which. Though such an idea of closeness is clearly and fundamentally relative, it is very difficult to define or evaluate precisely. That is, such relativity essence can't be directly measured with the presence of two point-clouds, but it can be somehow reflected only in settings involving three or more point-clouds. From this perspective, all existing distance measures commonly suffer from missing the data-clouds' essential senses of relative closeness locally and globally. For instance, recently Gromov-Wasserstein distance via Optimal Transport has been proposed as a direct evaluation of distance between two point-clouds \cite{13}. But it suffers from the known difficulty in handling high dimensionality (large $K$). So this distance measure likely misses the proper senses of relative closeness among point-clouds, especially when $K$ is big.

In this paper, we propose a simple computing approach to capture the relative closeness among all involving point-clouds without directly and explicitly evaluating pairwise cloud-to-cloud distance. The key idea is visible as follows: through randomly sampling a triplet of singletons from any triplet of point-clouds, we extract three partial ordering among the three pairs of cloud-to-cloud closeness. By taking one partial ordering as one win-and-loss in a tournament involving $\binom{L}{2}$ teams, we can build a dominance matrix that leads to a natural label embedding tree as a manifestation of heterogeneity on the global scale. Such a triplet-based brick-by-brick construction for piecing together a label embedding tree seems intuitive and natural. Indeed such a model-free approach is brand new to M.L. literature \cite{7,14}. The existing hierarchical methods build a somehow symbolic label embedding tree by employing a bifurcating scheme that nearly completely ignores the notion of heterogeneity \cite{7,15,16}.

After building a label embedding tree on the space of $L$ labels, we further derive a predictive graph, which is a weighted network with precisely evaluated linkages. This graph offers the detailed closeness from the predictive perspective as another key aspect of geometric information content of MCC.

To further discover the fine scale information content of MCC, we look into heterogeneity embraced by each label. Clustering analysis is applied on each label's point-cloud to bring out a natural clustering composition, and then label each cluster pertaining to a sublabel. By doing so across all labels, we result in a space of sublabel with much larger size than $L$. Likewise we compute a sublabel embedding tree and its corresponding predictive graph. These two geometries then constitute and represent the fine scale information content of MCC.

A real database, Major League Baseball (MLB) $PITCHf/x$, is analyzed for the purpose of application. Since 2008 the PITCHf/x database of MLB has been recording each every single pitch delivered by MLB pitchers in all games at its 30 stadiums. A record of a pitch is a measurement vector of 21 features. A healthy MLB pitcher typically pitches around 3000 pitches, which are algorithmically categorized into one of pitch-types:  Fastball, Slider, Change-up, curveball and others types.

We collect data from 14 $(=L)$ MLB pitchers, who threw around 1000 Fastball or more during the 2017 season. As one pitcher is taken as a label, his seasonal fastball collection is a point-cloud. It is noted that each pitcher tunes his Fastball slightly and distinctively when facing different batters under different circumstances of game. That is, multi-scale heterogeneity is inherently embedded into each point-cloud.

A potential feature set is selected based on permutation-based feature importance measure. The importance score is defined as the reduction in the performance of Random Forest after permuting the feature values. All real data illustrations for the entire computational developments throughout this paper is done with respect to a feature set consisting of 3 features: horizontal and vertical coordinates, and horizontal speed of a pitch at the releasing point. Results on two larger feature-set are also reported.

\begin{figure*}[h]
  \centering
  \includegraphics[scale=0.4]{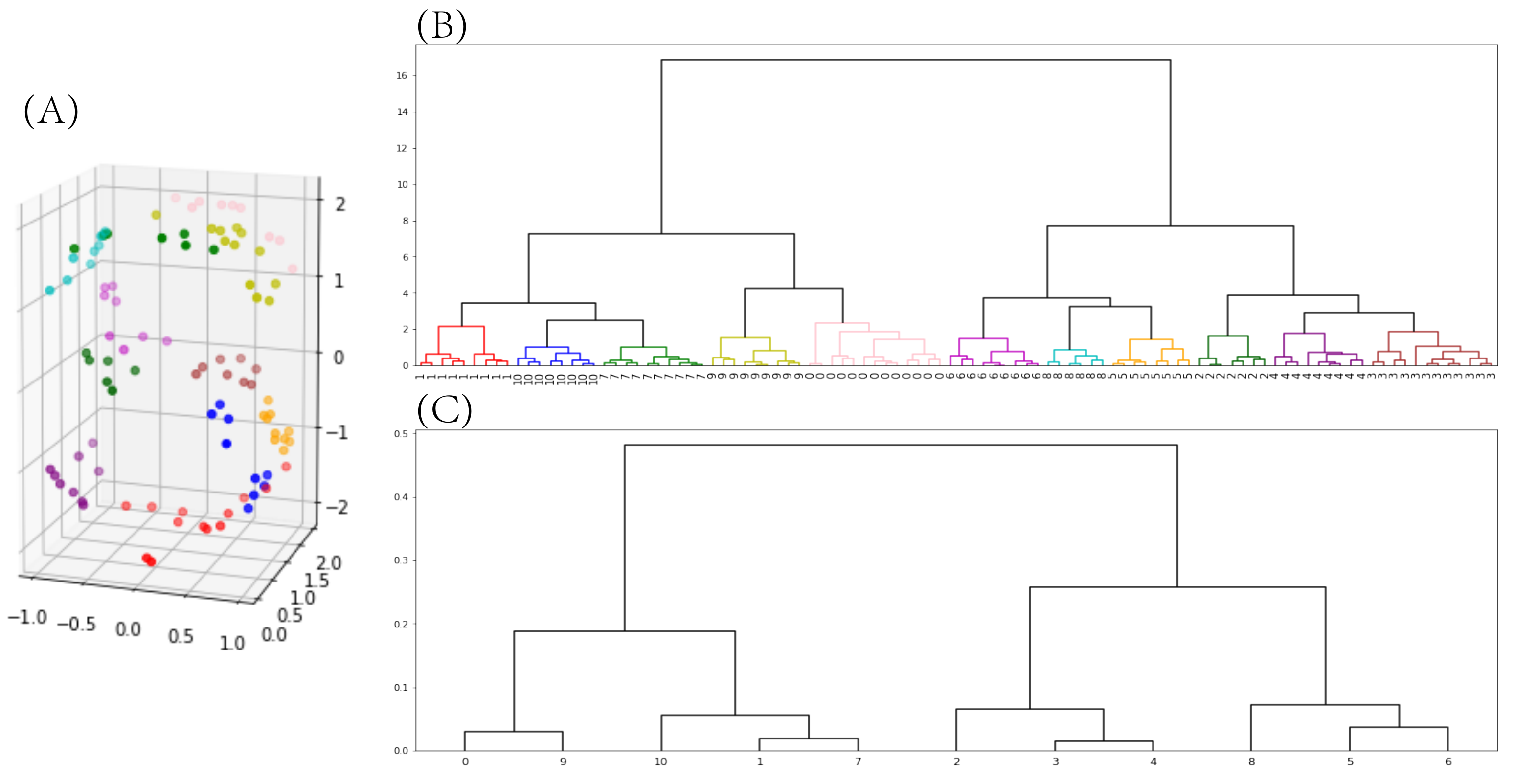}
  \caption{Illustrating example for the Algorithm of label embedding tree. (A) the 3D scatter plot of data; (B) the 11 labeled data-clouds defined by a tree; (C) the embedding tree.}
  \label{fig:illustration}
\end{figure*}

\section{A label embedding tree built by partial ordering.}
We develop a computing paradigm based on unsupervised machine learning to nonparametrically construct the label- and sub-label embedding trees in this paper. This paradigm is designed to be scalable to the three factors: $L$, $K$ and $N$. With a label-triplet, say $(La, Lb, Lc)$, in the brick-by-brick construction, partial ordinal relations are referred to: $D(La, Lb) < D(La, Lc)$ for example, where $D(.,.)$ is the unspecified ``distance'' between two label clouds. {\bf It is emphasized that the {\bf Alg.1} is devised to extract such relations without explicitly computing the three pairwise distances $D(.,.)$}.  These relations found among three point-clouds are stochastic in nature.

Given a triplet of labels $La,Lb,Lc$, if we randomly sample three singleton vectors in $R^K$, say $X_{La}$, $X_{Lb}$ and $X_{Lc}$: one from each of three labels, separately. A piece of information of partial ordering within the triplet can be shed by inequalities among Euclidian distances $d(.,.)$ among 3 singletons $X_{La}$, $X_{Lb}$ and $X_{Lc}$. That is, inequality $d(x_{La},x_{Lb})<d(x_{La},x_{Lc})$ provides a small piece of information about Labels $La$ and $Lb$ being closer than $La$ to $Lc$ and $Lb$ to $Lc$. By iteratively randomly sampling vector-triplets for a large number of times, say $T$, the probability of this relative closeness between $La$ and $Lb$ can be estimated as $\hat{P}(D(La,Lb)<D(La,Lc))=\sum_t{1_{D(La,Lb)<D(La,Lc)}}/T$.
Via Law of Large number, we arrive at the relative closeness information by aggregating partial ordering among all possible combination of three labels. Let $H$ be a square dominant matrix with $\binom{L}{2}=L(L-1)/2$ rows and columns. Each entry of $H$ records a probability that ``this unspecified distance $D(.,.)$ of a label-pair'' is dominated by the same unspecified distance of another label-pair. Denote $i_{xy}$ is the index of a label pair $Lx$ and $Ly$. The entry of $H$ in the $i_{ab}$th row and the $i_{cd}$th column records the related probability between these two label pairs.

\begin{equation}
H[i_{ab},i_{cd}]=P(D(La,Lb)<D(Lc,Ld))
\end{equation}

\noindent It is noted that $H(i,j)+H(j,i)$ is equal to $1$. In this way, $H$ realizes the partial ordering among all pairs of labels.

It is worth noting that such a dissimilarity matrix $\Bar{D}$ is by no means a metric satisfying triangular inequality or other properties. Here we illustrate the validity of this algorithm through a small example, as shown in Fig.\ref{fig:illustration}. A S-shape data set is simulated in $R^3$ space, see panel (A). Hierarchical clustering is implemented and a dendrogram is shown in panel (B). 11 clusters are obtained by cutting the dendrogram at a certain tree height and each cluster is marked with different color. Consider each cluster as a label, and a label embedding tree is created via Algorithm1 to show the hierarchical structure among those 11 classes in (C). It shows that our labeling tree built by only using partial ordering can reflect the original hierarchy among labels very well. In short, our dissimilarity matrix makes more sense in showing the natural label-cloud hierarchical dependency, which is the most advantage to distinguish our labeling tree from others.

There is a natural way to do classification based on this triplet partial ordering. We can simply assign a singleton or a batch of unlabeled sample with a new label $L_{new}$, which never appears in the previous label set. So there is supposed to be $L+1$ labels in total. Then, the triplet-version dissimilarity $\binom{L+1}{2} \times \binom{L+1}{2}$ matrix $H_{new}$ can be calculated for all those $L+1$ labels. The classified label is just the one that is the closest to the new label, see $\textbf{Alg.2}$. Actually, given the previous $\binom{L}{2} \times \binom{L}{2}$ matrix $H$ pre-trained, it is only necessary to calculate the rest $\binom{L+1}{2} \times L$ sub-matrix. That is to say, we randomly sample two singletons $X_{La}$ and $X_{Lb}$ from two labels $La$ and $Lb$, respectively, and sample one unlabeled sample $X_{new}$ from $L_{new}$. The partial ordering now turns out to compare $d(X_{La},X_{new})$ and $d(X_{Lb},X_{new})$. Via a large number of sampling, we gain information about $P(D(La, L_{new})<D(Lb, L_{new}))$ and its counterpart. Let $H_{new}$ record all newly added probabilities of such dominance. Then the label-pairwised dissimilarity matrix is calculated via the column sum of $H_{new}$. Therefore, the classification procedure is equivalent to aggregating all binary classifiers and vote according to the sum of probability, which is exactly one-versus-one classification with a soft vote strategy. One brand new property is that, when $L_{new}$ represents a unlabeled data-cloud, the geometry of this data-cloud is fully used in this predictive decision-making.
\bigbreak
\noindent\rule{12.5cm}{0.8pt}\\
\textbf{Alg.1} Label Embedding Tree\\
\rule{12.5cm}{0.4pt}\\
%\textbf{Input:} sample $x$'s and labels $y$'s\\
%\textbf{Denote:}\\
%$i:=$ index of label pair $(l_1,l_2)$\\
%$j:=$ index of label pair $(l_1,l_3)$\\
%$k:=$ index of label pair $(l_2,l_3)$\\
%$H(i,j)= P\{d(l_1,l_2)>d(l_2,l_3)\}$\\ $H(i,k)=P\{ d(l_1,l_2)>d(l_2,l_3)\}$\\ $H(j,k)=P\{ d(l_1,l_3)>d(l_2,l_3)\}$\\
%Note $H(i,j)+H(j,i)$ is not necessarily equal to $1$.\\
\textbf{Denote:} H is a $\binom{L}{2} \times \binom{L}{2}$ ranking matrix,\\
\hspace*{24pt}$H[i_{ab},i_{cd}]= P(D(La,Lb)<D(Lc,Ld))$\\
where $i_{ab}$ is the index of label pair $La$ and $Lb$,
$D(La,Lb)$ is their dissimilarity which is inaccessible.\\
\textbf{Initialize:} H with all entries 0 \\
\textbf{for} $(La,Lb,Lc)$ in all unique label triplets:\\
\hspace*{24pt}Randomly sampling a triplet of data for $T$ times with replacement, denoted\\
\hspace*{24pt}as $(X_{La}^{(1)},X_{Lb}^{(1)},X_{Lc}^{(1)})$, $(X_{La}^{(2)},X_{Lb}^{(2)},X_{Lc}^{(2)})$, ..., $(X_{La}^{(T)},X_{Lb}^{(T)},X_{Lc}^{(T)})$ \\
\hspace*{24pt}where $X_{L}$ is a single sample of data with label $y=L$\\
\hspace*{24pt}\textbf{for} $t$ in $1,...,T$:\\
\hspace*{24pt}\hspace*{24pt}\textbf{if} $d(X_{La}^{(t)},X_{Lb}^{(t)})<d(X_{La}^{(t)},X_{Lc}^{(t)})$: $H[i_{ab},i_{ac}]+=1/T$\\
\hspace*{24pt}\hspace*{24pt}\textbf{else} $H[i_{ac},i_{ab}]+=1/T$\\
\hspace*{24pt}\hspace*{24pt}\textbf{if} $d(X_{La}^{(t)},X_{Lb}^{(t)})<d(X_{Lb}^{(t)},X_{Lc}^{(t)})$: $H[i_{ab},i_{bc}]+=1/T$\\
\hspace*{24pt}\hspace*{24pt}\textbf{else} $H[i_{bc},i_{ab}]+=1/T$\\
\hspace*{24pt}\hspace*{24pt}\textbf{if} $d(X_{La}^{(t)},X_{Lc}^{(t)})<d(X_{Lb}^{(t)},X_{Lc}^{(t)})$: $H[i_{ac},i_{bc}]+=1/T$\\
\hspace*{24pt}\hspace*{24pt}\textbf{else} $H[i_{bc},i_{ac}]+=1/T$\\
\hspace*{24pt}\textbf{end for}\\
\textbf{end for}\\
Calculate $K \times K$ labeling dissimilarity matrix $\Bar{D}$\\
$\Bar{D}(La,Lb)=E_{Lx,Ly}\{P(D(Lx,Ly)<D(La,Lb))\}=\sum_{j}{H(j,i_{ab})}/\binom{L}{2}$\\
\textbf{Output:} a hierarchical clustering tree based on the dissimilarity matrix $\Bar{D}$\\
\rule{12.5cm}{0.8pt}
\bigbreak
\noindent We can also sample $X_{La}$ and $X_{Lb}$ from the neighbors of $X_{new}$ to extract the partial ordering locally. Let's choose M-nearest neighbors of $X_{new}$ constrained in the data with label $La$, denoted as $X_{M|La}=(X_{La}^{(1)},X_{La}^{(2)},...,X_{La}^{(M)})$, and so is $X_{M|Lb}$. We look at whether there are relatively more $La$'s compared with $Lb$'s in the M nearest neighbors. This classification becomes k-Nearest Neighbor with tuning parameter $k$ chosen to be $M$. If we repeat the aforementioned procedure for a large number of times, we have another way of extracting information of $P(D(La, L_{new})<D(Lb, L_{new}))$. Thus, Algorithm2 is equivalent to one-versus-one classification with k-NN as its classifier. These properties also explain why our triplet comparison is so important.

\begin{figure}[h]
  \centering
  \includegraphics[scale=0.38]{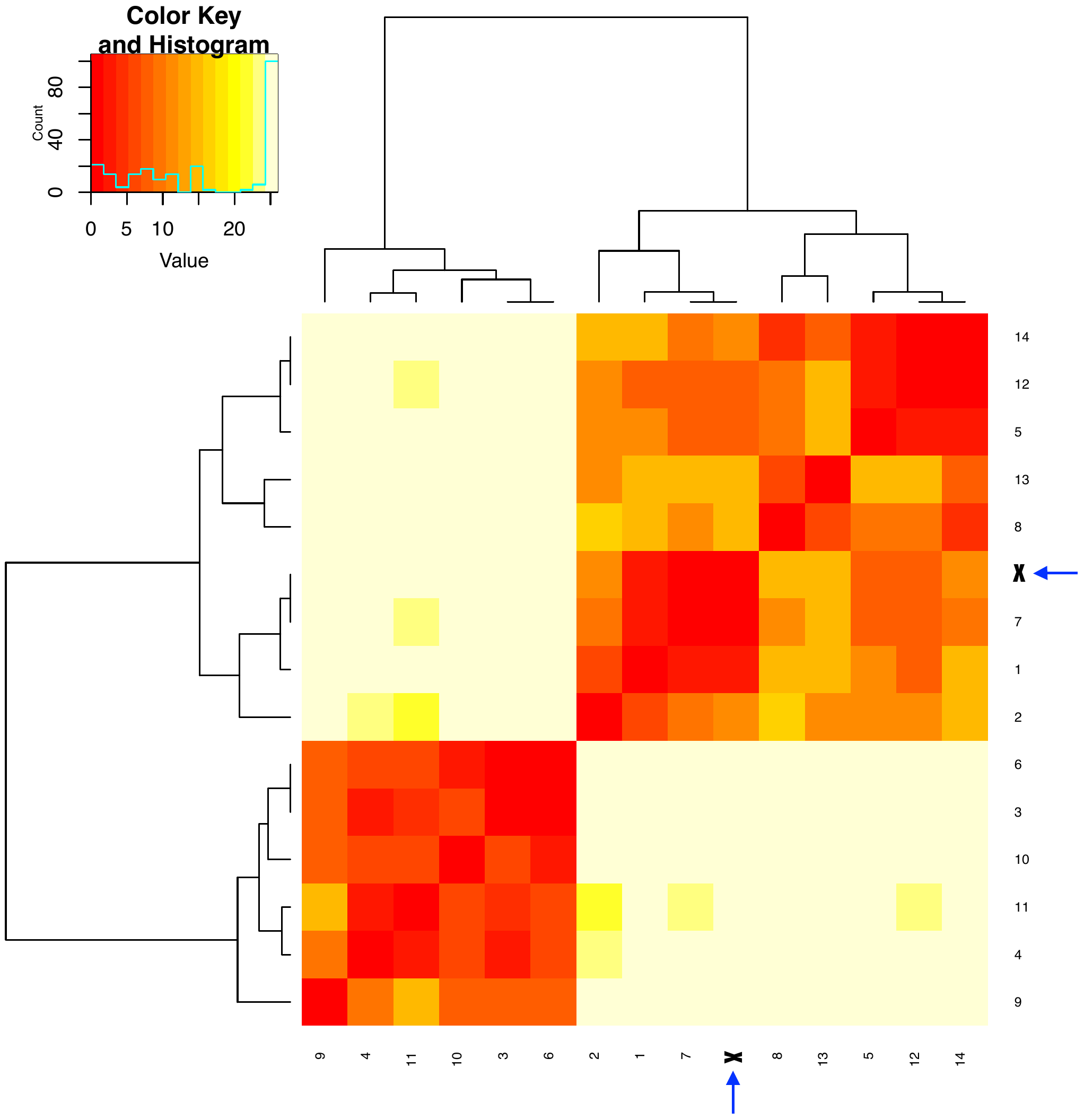}
  \caption{Label embedding tree of 14 pitchers with a heatmap of ``distance'' derived from a computed $H$ and an illustrating example of classifying an unknown label $\textbf{X}$; the ground truth is label $7$}
  \label{fig:tree_location}
\end{figure}

Besides, Algorithm2 can indicate where the unknown label is located within the previous label embedding tree. The label embedding tree with an unknown label embedded is clear to view which labels are mixed with the unknown label in a small branch and which labels is far away. See Fig.\ref{fig:tree_location} for an illustration.

\noindent The number of sampling iteration $T$ is supposed to be as large as possible. In practice, $T$ should be chosen dependent on the sample size of each label. If the data is balanced, $T=N/L$, otherwise, $T=max_{i}N_i$ to cover the biggest label data cloud, where $N_i$ is the sample size for label $i$. So the time complexity is $O(NKL^2)$.

When $L$ is small or moderate, consider a setting with the number of all possible triplets, $\binom{L}{3}$, being not overwhelmingly big. We perform Algorithm1 on all possible triplets to fill up the $\binom{L}{2} \times \binom{L}{2}$ dominance matrix, $H$. Each of column sum of $H$ tells how many times a label-pair's distance is dominated by distances of all other pairs. So the bigger a column sum is, the larger degree of similarity of this label pair is. Therefore the $\binom{L}{2}$-vector of column sums of $H$ can be transformed into a natural $L \times L$ similarity matrix, say $\Bar{S}$, among all involving labels. In contrast, the $\binom{L}{2}$-vector of row sums of $H$ is a distance (dissimilarity) matrix, say $\Bar{D}$, of all labels. Such a $\Bar{S}$ or $\Bar{D}$ will afford a hierarchy, which is the label embedding tree.
\newpage
\noindent\rule{12.5cm}{0.8pt}\\
\textbf{Alg.2:} Classify $X_{new}$ with an unknown label $Lx$\\
\rule{12.5cm}{0.4pt}\\
\textbf{Input:} a $\binom{L}{2} \times \binom{L}{2}$ matrix $H$ obtained from \textbf{Alg.1}\\
\textbf{Initialize:} a $\binom{L+1}{2} \times \binom{L+1}{2}$ ranking matrix $H_{new}$\\
$H_{new}[1:\binom{L}{2},1:\binom{L}{2}]=H$ and the rest entry 0.\\
\textbf{for} $(La,Lb)$ in all unique label pairs:\\
\hspace*{24pt}Randomly sampling a pair of data for $T$ times with replacement, and \\
\hspace*{24pt}concatenate it with $X_{new}$ to make a triplet, denoted as $(X_{La}^{(1)},X_{Lb}^{(1)},X_{new})$, \hspace*{24pt}$(X_{La}^{(2)},X_{Lb}^{(2)},X_{new})$, ..., $(X_{La}^{(T)},X_{Lb}^{(T)},X_{new})$ \\
\hspace*{24pt}where $X_{L}$ is a single sample of data with label $y=L$\\
\hspace*{24pt}\textbf{for} $t$ in $1,...,T$:\\
\hspace*{24pt}\hspace*{24pt}\textbf{if} $d(X_{La}^{(t)},X_{new})<d(X_{Lb}^{(t)},X_{new})$,
$H_{new}[i_{ax},i_{bx}]+=1/T$\\
\hspace*{24pt}\hspace*{24pt}\textbf{else} $H_{new}[i_{bx},i_{ax}]+=1/T$\\
\hspace*{24pt}\hspace*{24pt}where $i_{ax}$ and $i_{bx}$ are indices for label pair $(La,Lx)$ and $(Lb,Lx)$\\
\hspace*{24pt}\textbf{end for}\\
\textbf{end for}\\
\textbf{Output1:} Classification result $a^*$, if\\ \hspace*{24pt}$i^*=i_{a^*x}$, $i^*=\argmin_{i} \sum_{j}{H(j,i)}$\\
Get $(L+1) \times (L+1)$ dissimilarity matrix $\Bar{D}_{new}$\\
$\Bar{D}_{new}(La,Lb)=\sum_{j}{H(j,i_{ab})}/\binom{L+1}{2}$\\
\textbf{Output2:} a hierarchical clustering tree on $\Bar{D}_{new}$\\
The tree can also return in which branch the unknown label $Lx$ locates from the previous labeling tree.\\
\rule{12.5cm}{0.8pt}\\

When it is too expansive to compute a full version of $H$, then we start with a sparse version, says $H'$. By applying the transitivity property in dominance relationship, we can resolve the sparsity issue by making product matrix like $H=H' \times H'$ to record all indirect dominance with one intermediate \cite{17}, see the Algorithm-A in Appendix. By embracing such transitivity, as confirmed in our experiment, a reliable distance dominance matrix $H$ can be resulted.

\begin{figure*}[h]
  \centering
  \includegraphics[scale=0.4]{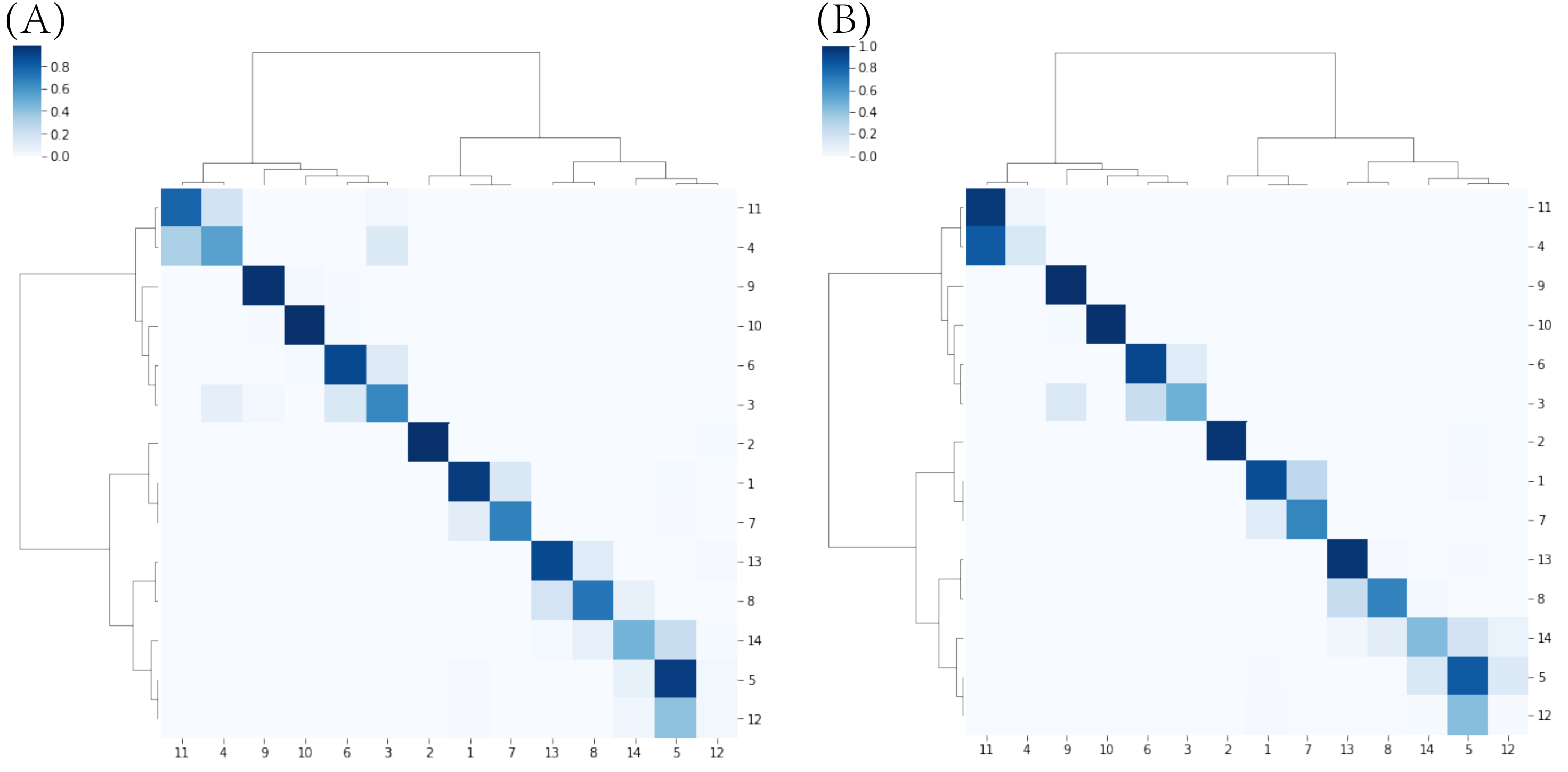}
  \caption{Label embedding tree superimposed on its confusion matrix: (A) Classification being driven to the tree bottom with a singleton label candidate; (B) Classification can stop early at a tree inter-node.}
  \label{fig:14pitcherHM}
\end{figure*}

\section{Tree-descent schedule and error flow}
With a label embedding tree, a very efficient decision-making process can be devised via tree descend framework as depicted in {\bf Alg.3}. This algorithm works for any bi-class classifier by making a chain of decisions from top-to-bottom levels of the label embedding tree. So our label embedding tree becomes a scalable platform for decision-making with respect to the number of labels ($L$). In fact the tree somehow provides an ideal setting for distance metric learning \cite{18}, because similar labels have been clustered together.

For prediction purpose, ideally the tree's binary branching structure can allow us to arrive at a singleton label at the bottom of tree, or a small set of label as a small tree branch by avoiding any risk of making any major mistake. A threshold $\theta$ defined in Algorithm3 works for risk control. If the probability of classification is less than $\theta$, say 0.8, we have less confidence to descend the labeling tree further, so early stop the iteration and return a label set. This fact can be visualized from our construction of predictive graph below.
\bigbreak
\noindent\rule{12.5cm}{0.8pt}\\
\textbf{Alg.3} Classify $X_{new}$ via descending label embedded tree with an early stop\\
\rule{12.5cm}{0.4pt}\\
\textbf{Input:} a label embedding tree $B$; a trained Binary Classifier $F$; a threshold $\theta$ to stop descending tree\\
\textbf{Denote:}\\
$B_{Left}$ and $B_{Right}$ are the left and right branch on the root node of tree $B$\\
$F_{L}(X_{new})$ returns the probability of classifying $X_{new}$ into Left branch\\
$F_{R}(X_{new})$ returns the probability of classifying $X_{new}$ into Right branch\\
\textbf{while} $(|B|>1$ \& $max\{F_{L}(X_{new}),F_{R}(X_{new})\}>\theta):$\\
\hspace*{24pt}\textbf{if} $F_{L}(X_{new})>F_{R}(X_{new})$, \textbf{then}
$B \leftarrow B_{Left}$\\
\hspace*{24pt}\textbf{else} $B \leftarrow B_{Right}$\\
\textbf{end while}\\
\textbf{Output:} label(s) under the current tree $B$\\
\rule{12.5cm}{0.8pt}
\bigbreak
Let ${\cal Y}=\{L_j\}^L_{1}$ and ${\cal F}=\{f_i\}^K _{1}$ be the ensembles of label and feature, respectively. Denote a computed label embedding tree as ${\cal B}[{\cal F}]$. We derive a label predictive graph, denoted by ${\cal G}[{\cal F}]$, based on a confusion matrix. All classification results are collectively summarized into an asymmetric error-flow matrix ${\cal E}[{\cal F}]=[e_{i,j}]$ with directed error-flows $(e_{i,j}, e_{j,i})$ between any label pair $(L_i, L_j)$ are the percentages of wrong decisions by predicting $L_i$ to be $L_j$, and vice versa. ${\cal G}[{\cal F}]$ is a weighted network or graphic representation of ${\cal E}[{\cal F}]$, see Fig.\ref{fig:14pitcherHM} for two predictive graphs of 14 MLB pitcher-labels.

The essence of ${\cal G}[{\cal F}]$ is that its pairwise directional links $\{ (e_{i,j}, e_{j,i})\}$ realistically reflects unequal mixing configurations of labels $L_i$ from $L_j$. The utility of ${\cal G}[{\cal F}]$ is that it allows a smallest predictive label set, while achieving a nearly perfect precision. Such an asymmetry, See Fig.\ref{fig:14pitcherHM}, is invaluable in understanding the MCC setting and in explaining decision-making. This perspective is completely lost when a direct distance measure is forcefully employed.

\begin{figure*}[h]
  \centering
  \includegraphics[scale=0.3]{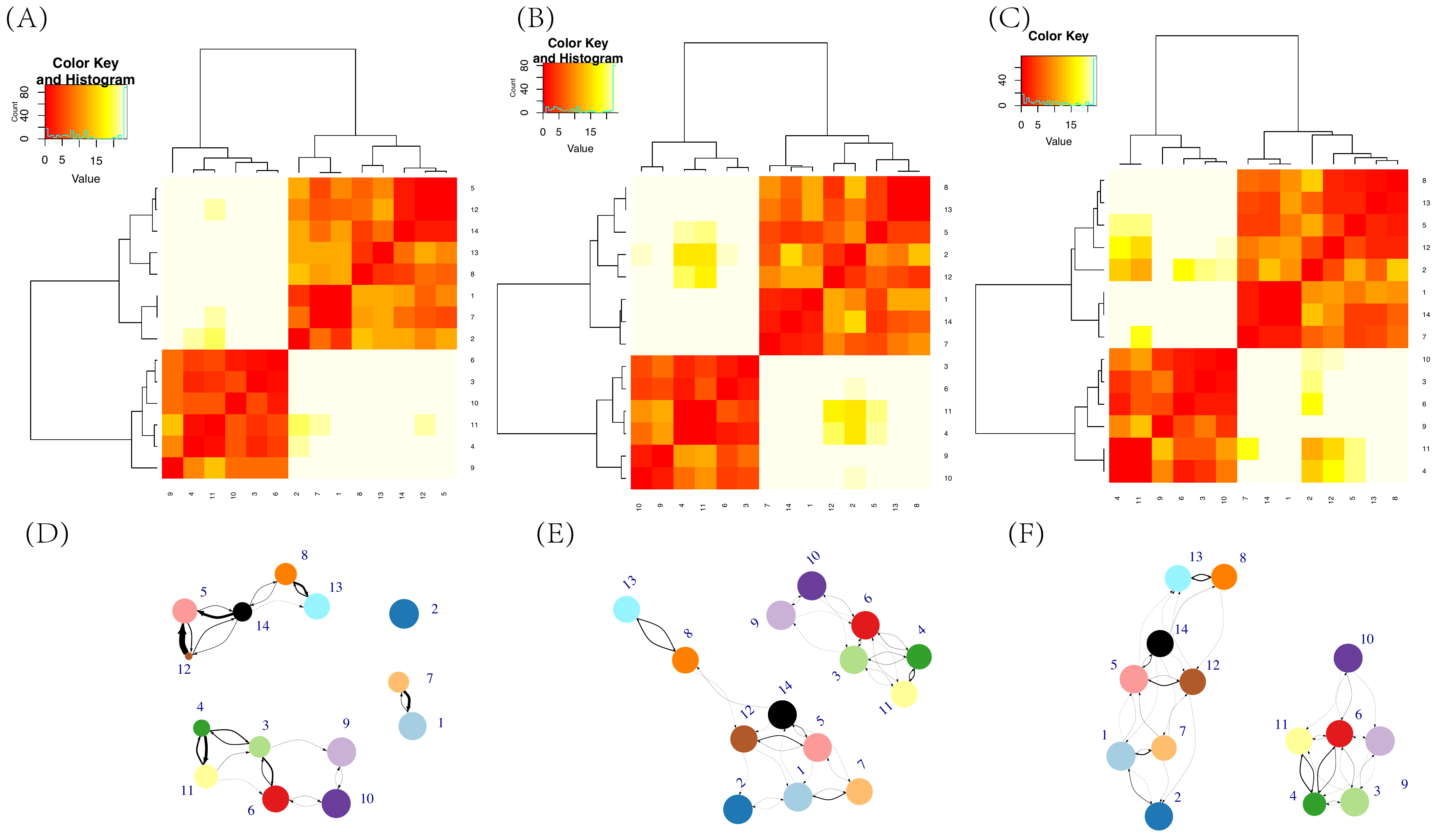}
  \caption{dissimilarity matrix and predictive graphs calculated on 3 different Feature Groups with increasing sizes (see Group 1, 3 and 4 in Appendix)). (A),(B),(C) illustrate the dissimilarity matrix with a label embedding tree embedded on the row and column axis. The label number is the index of a baseball pitcher. There are 14 different pitcher, labeled from 1 to 14; (D),(E),(F) are predictive graphs that visualize the bi-class cut tree descending result.}
  \label{fig:graphG1}
\end{figure*}

%\begin{figure*}[h]
%  \centering
%  \includegraphics[scale=0.35]{Graph_GB.pdf}
%  \caption{label embedding tree and predictive graphs calculated on Feature Group3. (A) illustrate the dissimilarity matrix with a label embedding tree embedded on the row and column axis. The label number is the index of a baseball pitcher; both (B) and (C) are predictive graphs that visualize the tree descending result. (B) is based on k-NN, while (C) is on PLR (see Algorithm-B in Appendix.}
% \label{fig:graphGB}
%\end{figure*}

%\begin{figure*}[h]
%  \centering
%  \includegraphics[scale=0.35]{Graph_All.pdf}
%  \caption{label embedding tree and predictive graphs calculated on All Features. (A) illustrate the dissimilarity matrix with a label embedding tree embedded on the row and column axis. The label number is the index of a baseball pitcher; both (B) and (C) are predictive graphs that visualize the tree descending result. (B) is based on k-NN, while (C) is on PLR.}
%\label{fig:graphGAll}
%\end{figure*}

%\subsection{Data-driven intelligence for explainable decision-making}
With the two explicit and visible geometries embraced by the computed label embedding tree and its corresponding predictive graph as the MCC information content with respect to feature set ${\cal F}$, the linkages between the label space ${\cal Y}$ and the collection of point-clouds defined by feature set ${\cal F}$ become evidently explainable. %Thus, the collective of linkages based knowledge is called Data-driven Intelligence (D.I.). In sharp contrast with artificial Intelligence (A.I.), such explainable D.I. is free from any risk due to artificial modeling structures and assumptions, so it.
It is clear to see that the predictive graph is possible to guide us to error-free decision-making if our decision is in a form of a set of potential label candidates, rather than restricted to a singleton. This fact leads us to reflect on the common phenomenal issue: Why predicting an unlabeled singleton has to be prone to error? There are at least two key reasons. First, a predictive object can be caught deep within some point-clouds of wrong labels, not just the right one. Therefore, involving all labels' data-clouds at once for such prediction is not ideal. To ameliorate such a situation, a decision-making process descents from the top of a label embedding tree is strategic since MCC's information content is fully used. The second reason is that we ignore what amount of information is available, and simultaneously force ourselves to make a single pick of label.

\section{Fine scale information content of MCC.}
It is known that each label's point-cloud contains its own label specific heterogeneity. Discovering and accommodating such heterogeneity into MCC's information content in a collective fashion is another essential part of our data-driven computational endeavors. Since our label embedding tree can represent the natural hierarchical structure among separated data clouds, it is straightforward for us to decompose one label's point cloud into separate sublabel clusters and then implement Algorithm1. The sublabel clusters are empirically discovered from each label through a hierarchical clustering tree built upon this label's point could. On the MLB pitching MCC setting, 139 sublabels are generated. We then likewise construct a sublabel embedding tree and its corresponding $139\times 139$ confusion matrix.

\begin{figure*}[!h]
  \centering
  \includegraphics[scale=0.33]{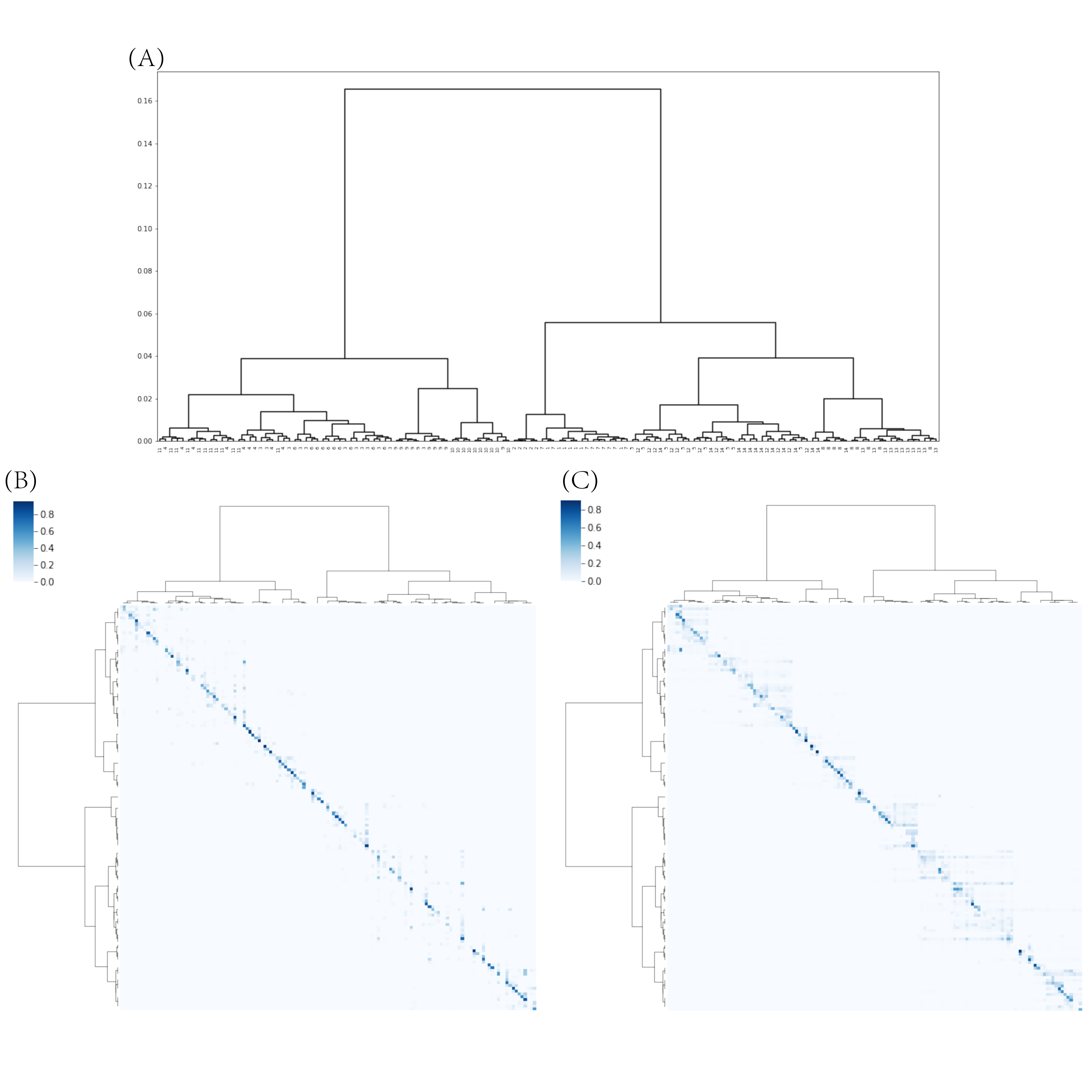}
  \caption{Fine scale multiscale geometry of 139 sublabels, which belong to 14 pitchers labeled from 1 to 14. (A)The sublabel embedding tree; (B)the confusion matrix with a singleton label candidate; (C) predictions stop early at a tree inter-node.}
  \label{fig:139subpitcherHM}
\end{figure*}

Both geometries of fine scale MCC's information content are shown in the three panels of Fig.\ref{fig:139subpitcherHM}. They explicitly reveal detail and complex mixing patterns among the 139 sublabel specific point-clouds. Such fine scale information to a great degree reflect the coarse scale information, but at the same time shed many new lights on its own. For instance, we see how diverse subtypes are belonging to a pitcher's fastball. If all his subtypes are located in a relative small branch of the sublabel embedding tree, then this pitcher fastball pitches are rather uniform. In contrast, if his subtypes are located across several far apart branches, then this pitcher's fastball pitches are difficult to predict. Further we examine in explicit detail how his subtypes are mixing with other pitchers' via a predictive graph. Such examinations allow us to discover how and why this pitcher is in common with which pitchers, and how and why he is distinct with which pitchers. That is, these two geometries are platforms for discovering and establishing many ways of comparing MLB pitchers from many aspects. All these discoveries as diverse parts of the collective knowledge made possible by the fine scale of information content of MCC.

\section{Conclusion}
The coarse and fine scales of information contents of MCC afford us to zoom-in and zoom-out to discover Data-driven Intelligence (D.I.) in visible and explainable fashion. The implied nearly perfect decision-making allows researchers to be responsible. We hope such a D.I. mindset can prevail from sciences to health industries, and beyond. Promoting D.I. is same as promoting truth and knowledge already contained in data. Human might have been very wasteful in casting away invaluable knowledge by only focusing on forceful prediction.

Finally we make a remark on feature selection. Our standpoint here is that perfect decision-making is the prerequisite on any prediction issue occurring in sciences and health industries. Over these fields, any prediction needs to rightly reflect the amount of information available from data. At the same time, all decision-makers have to be responsible on what they decide. Their subject-matter sensitive criteria can be easily based on the two geometries of MCC's information content. That is, the task of feature selection shall be based on the ${\cal F}$ and be subject-matter sensitive. Such a standpoint is illustrated in Fig.\ref{fig:graphG1}. By comparing the three sets of geometric information contents pertaining to three feature-sets (feature information given in Appendix), we gain different understanding and knowledge regarding the 14 MLB pitchers. We explain such Data-driven Intelligence(D.I.) pertaining to different sets of feature. That is why a prediction is better feature-set sensitive.

In summary, at least under MCC settings, Data-driven Intelligence is one basic principle objective of machine learning in Data Science as well as in Artificial Intelligence.

%%%%%%%%%% Insert bibliography here %%%%%%%%%%%%%%


%------------------------------------------------

%\bibliographystyle{abbrv}
\bibliographystyle{splncs04}
\bibliography{egbib.bib}

\begin{thebibliography}{10}
\providecommand{\url}[1]{\texttt{#1}}
\providecommand{\urlprefix}{URL }
\providecommand{\doi}[1]{https://doi.org/#1}

\bibitem{15}
{Aixin Sun}, {Ee-Peng Lim}: Hierarchical text classification and evaluation.
  In: Proceedings 2001 IEEE International Conference on Data Mining. pp.
  521--528 (2001). \doi{10.1109/ICDM.2001.989560}

\bibitem{10}
Allwein, E.L., Schapire, R.E., Singer, Y.: Reducing multiclass to binary: a
  unifying approach for margin classifiers. Journal of Machine Learning
  Research  \textbf{1},  113--141 (2000)

\bibitem{6}
Amit, Y., Fink, M., Srebro, N., Ullman, S.: Uncovering shared structures in
  multiclass classification. In: Proceedings of the 24th International
  Conference on Machine Learning. p. 17–24. ICML '07, Association for
  Computing Machinery, New York, NY, USA (2007). \doi{10.1145/1273496.1273499},
  \url{https://doi.org/10.1145/1273496.1273499}

\bibitem{7}
Bengio, S., Weston, J., Grangier, D.: Label embedding trees for large
  multi-class tasks. In: Proceedings of the 23rd International Conference on
  Neural Information Processing Systems - Volume 1. p. 163–171. NIPS'10,
  Curran Associates Inc., Red Hook, NY, USA (2010)

\bibitem{14}
Bhatia, K., Jain, H., Kar, P., Varma, M., Jain, P.: Sparse local embeddings for
  extreme multi-label classification. In: Proceedings of the 28th International
  Conference on Neural Information Processing Systems - Volume 1. p. 730–738.
  NIPS'15, MIT Press, Cambridge, MA, USA (2015)

\bibitem{12}
Cisse, M.M.: Efficient Extreme Classification. Data Structures and Algorithms.
  Ph.D. thesis, Université Pierre et Marie Curie - Paris VI (2014),
  https://tel.archives-ouvertes.fr/tel-01142046/document

\bibitem{9}
Ciss\'{e}, M., Usunier, N., Artieres, T., Gallinari, P.: Robust bloom filters
  for large multilabel classification tasks. In: Proceedings of the 26th
  International Conference on Neural Information Processing Systems - Volume 2.
  p. 1851–1859. NIPS'13, Curran Associates Inc., Red Hook, NY, USA (2013)

\bibitem{3}
Deng, J., Berg, A.C., Li, K., Fei-Fei, L.: What does classifying more than
  10,000 image categories tell us? In: Computer Vision -- ECCV 2010. pp.
  71--84. Springer Berlin Heidelberg, Berlin, Heidelberg (2010)

\bibitem{17}
Fushing, H., Fujii, K.: Mimicking directed binary network for exploring
  systemic sensitivity: Is ncaa fbs a fragile competition system. Frontiers in
  Applied Mathematics and Statistics  \textbf{2}, ~9 (2016)

\bibitem{8}
Gupta, M.R., Bengio, S., Weston, J.: Training highly multiclass classifiers.
  Journal of Machine Learning Research  \textbf{15},  1461--1492 (2014)

\bibitem{11}
Hastie, T., Tibshirani, R.: Classification by pairwise coupling. The Annals of
  Statistics  \textbf{26}(2),  451--471 (2001)

\bibitem{16}
Kosmopoulos, A., Partalas, I., Gaussier, E., Paliouras, G., Androutsopoulos,
  I.: Evaluation measures for hierarchical classification: a unified view and
  novel approaches. Data Mining and Knowledge Discovery  \textbf{29},  820--865
  (2015)

\bibitem{2}
{Lecun}, Y., {Bottou}, L., {Bengio}, Y., {Haffner}, P.: Gradient-based learning
  applied to document recognition. Proceedings of the IEEE  \textbf{86}(11),
  2278--2324 (1998). \doi{10.1109/5.726791}

\bibitem{1}
Russell, S.J., Norvig, P.: Artificial Intelligence: A Modern Approach (3nd
  ed.). Prentice Hall, Upper Saddle River, NJ, USA (2009)

\bibitem{13}
Solomon, J.: Optimal transport on discrete domains  (2018),
  arxiv.org/abs/1801.07745

\bibitem{5}
Tsoumakas, G., Katakis, I., Vlahavas, I.: A review of multi-label
  classification methods. In: In Proceedings of the 2nd ADBIS Workshop on Data
  Mining and Knowledge Discovery (ADMKD 2006). pp. 99--109 (2006)

\bibitem{18}
Weinberger, K.Q., Saul, L.K.: Distance metric learning for large margin nearest
  neighbor classification. Journal of Machine Learning Research  \textbf{10},
  207--244 (2009)

\bibitem{4}
Weinberger, K.Q., Chapelle, O.: Large margin taxonomy embedding for document
  categorization. In: Advances in Neural Information Processing Systems 21,
  Proceedings of the Twenty-Second Annual Conference on Neural Information
  Processing Systems, Vancouver, British Columbia, Canada, December 8-11, 2008.
  pp. 1737--1744. Curran Associates, Inc. (2008)

\end{thebibliography}

\newpage

\section*{Appendix}

\subsection*{\bf Data Accessibility:} The pitching data is available in PITCHf/x database belonging to Major League Baseball via \url{http://gd2.mlb.com/components/game/mlb/}.

\subsection*{Feature explanation from PITCHf/x}
A pitched baseball flight captured by 20 pairs of images via a pair of 60Hz cameras, which have orthogonal optical axes and cover the field of view between pitcher's mound and home plate, are determined with respect to the field coordinates. These images and estimated coordinates are converted into 21 features to characterize the flight's aerodynamics.

The 21 features are briefly described as follows:

1.	The starting speed (``start speed'') is measured when the ball is at the point 50 fts away from the home plate, which is very close to the horizontal and vertical coordinates of release point $(x0, z0)$ of a pitch.

2.	The spin direction (``spin  dir'') is determined by assuming spin-axis being perpendicular to the movement direction, while spin rate (``spin   rate'') is the number of rotations per minute.

3.	Vertical and horizontal movement measurements, denoted by ``pfx-z'' and ``pfx-x'', respectively. Topspin and backspin cause positive and negative vertical movements ``pfx-z''. Therefore this feature has a high association with ``start  speed" for pitchers, who has the high speed fastball as his chief pitch-type in his repertoire, than for pitchers, who doesn't. The feature ``pfx-z'' is also associated with features related to how a baseball trajectory curves.

4.	A baseball trajectory from release point to the home plate is coupled with two straight lines: the tangent line at the release point $(x_0, z_0)$ and the line links the release point and the trajectory's end point. The angle between these two lines is termed ``break angle'', while the maximum distance between the baseball trajectory and the second straight line is called and denoted as ``break length''.  Therefore the three features: ``pfx-z'', ``break  angle'' and ``break length'', are highly associated with each other.

5.	The remaining features are three directions of speeds and accelerations at the release point, named ``vx0, vy0,vz0'' and ``ax, ay, az'', respectively, or play only auxiliary roles, like ``break y'', ``x'', and ``y''.
\\

Feature Group1:
``x0", ``z0", and ``vx0"
\\

Feature Group2:
``x0", ``z0", ``vx0", ``vy0", ``start-speed", ``end-speed", and ``spin-dir"\\

Feature Group3:
``x0", "z0", ``vx0", ``vy0", ``start-speed" ``end-speed", ``spin-dir", ``spin-rate", ``break-angle", ``pfx-x", and ``pfx-z"\\

Feature Group4: all 21 features

\bigbreak
\noindent\rule{12.5cm}{0.8pt}\\
\textbf{Alg.A} Label Embedding Tree (Sparse)\\
\rule{12.5cm}{0.4pt}\\
\textbf{Alg.1} is applied to get the dominance matrix $H^{'}$ with a smaller sampling iteration $T$\\
$H= H^{'} + H^{'} \times H^{'}$\\
$H(i,j)= min\{H(i,j),1\}$\\
$\Hat{D}(La,Lb)=\sum_{j}{H(j,i_{ab})}/\binom{L}{2}$\\
\textbf{Output:} a label embedding tree based on $\Hat{D}$\\
\rule{12.5cm}{0.8pt}

\end{document}